\documentclass[letterpaper, 10 pt, conference]{ieeeconf}  % Comment this line out if you need a4paper

\IEEEoverridecommandlockouts                            
\newcommand{\red}[1]{\textcolor{red}{#1}}

\usepackage{color}
\usepackage{amsmath}
\usepackage{amssymb,latexsym}
\usepackage[colorinlistoftodos, color=blue!20]
{todonotes}
\usepackage{tikz}
\usetikzlibrary{3d}

\title{\LARGE \bf
Robust Orientation Estimation with TRIAD-aided Manifold EKF
}
\author{Arjun Sadananda$^{1}$, Ravi Banavar$^{2}$ and Kavi Arya$^{3}$ 
\thanks{$^{1,2}$
        Centre for Systems and Control, Indian Institute of Technology Bombay, Mumbai, 400076, Maharashtra, India $^{3}$
        Department of Computer Science and Engineering, Indian Institute of Technology Bombay, Mumbai, 400076, Maharashtra, India $^1${\tt\small arjun.sadananda@gmail.com}, $^2${\tt\small banavar@iitb.ac.in}, $^3${\tt\small kavi@cse.iitb.ac.in}}
}
\usepackage{mymacros}
\usepackage{ragged2e} 
\usepackage{verbatim}
\usepackage{tabularx}
\usepackage{amsmath}
\usepackage{bm}
\usepackage{ulem}

\usepackage[lofdepth,lotdepth]{subfig}

\begin{document}
\maketitle
\thispagestyle{empty}
\pagestyle{plain}

\pagenumbering{gobble}

\newcommand{\RS}[1]{\sout{\red{#1}}}
%%%%%%%%%%%%%%%%%%%%%%%%%%%%%%%%%%%%%%%%%%%%%%%%%%%%%%%%%%%%%%%%%%%%%%%%%%%%%%%%
\begin{abstract}
The manifold extended Kalman filter (Manifold EKF) has found extensive application
for attitude determination. Magnetometers employed as sensors for such attitude
determination are easily prone to disturbances by their sensitivity to calibration and external magnetic fields. The TRIAD (Tri-Axial Attitude Determination) algorithm is well-known as a sub-optimal attitude estimator. In this article, 
we incorporate this sub-optimal feature of the TRIAD in mitigating the influence
of the magnetometer reading in the pitch and roll axis determination in the 
Manifold EKF algorithm. We substantiate our results with experiments.
\end{abstract}

%%%%%%%%%%%%%%%%%%%%%%%%%%%%%%%%%%%%%%%%%%%%%%%%%%%%%%%%%%%%%%%%%%%%%%%%%%%%%%%%
% \pagenumbering{roman}

\section{Introduction}

Accurate orientation estimation is critical for a wide range of applications, such as in Unmanned Aerial Vehicles (UAVs), mobile devices and robotics. Numerous studies have been dedicated to improving sensor orientation estimation. The Multiplicative Extended Kalman Filter \cite{lefferts1982kalman}, remains the popular choice for a vast majority of applications \cite{crassidis2007survey}. In this approach an “error-quaternion”  \cite{Markley2003} is defined, which is then transformed to a 3-vector. This vector is used to build the covariance matrix, and define a “3×3 representation of the quaternion covariance matrix”. However, there are details in this adaptation that are being developed, such as the “covariance correction step” \cite{mueller2017covariance}.

\cite{bernal2019kalman} presents a new viewpoint by noticing that unit quaternions live on a manifold $\mathsf{S}^3$ (the unit sphere in $\mathbb{R}^4$). This work uses basic concepts from manifold theory to deﬁne the mean and covariance matrix of a distribution of unit quaternions. With these deﬁnitions, an EKF-based estimator is developed, arriving at the concepts of “multiplicative update” and “covariance correction step” in a natural and satisfying way. Further, \cite{ge2023note} generalises this approach to building EKF on a manifold and also compares it with the naive coordinate implementation of the EKF. 
% One of the big reasons for the Kalman Filter's popularity is its simplicity and its flexibility to incorporate a great variety of measurements. 
In this work, we introduce a second reference vector, the magnetometer, along with the accelerometer, to the Manifold Extended Kalman Filter (Manifold EKF) estimator \cite{bernal2019kalman}. However, this presents new challenges, particularly due to the sensitivity of magnetometers to magnetic disturbances.

% The 3D orientation of a system is typically represented using Euler angles—roll, pitch, and yaw. Roll and pitch describe the system's attitude, while yaw corresponds to the heading. 
% Low-cost systems for orientation estimation often rely on light and cheap Micro-electromechanical Systems (MEMS) inertial and magnetic measurement units (IMMUs). 
% Accurate sensor 3D orientation estimation remains challenging, particularly due to magnetic disturbances affecting the magnetometer. 
In practical applications, such as Unmanned Aerial Vehicles (UAVs), roll and pitch are often estimated using only gyroscopes and accelerometers. However, this comes at the cost of gyro integration drift in the yaw axis, necessitating the use of magnetometers to correct for these drifts. Unfortunately, magnetometer measurements are often distorted by hard-iron and soft-iron magnetic effects \cite{Fan2018}, degrading the overall orientation estimation

While using both accelerometer and magnetometer inputs in the MEKF theoretically should improve yaw estimation, the method's performance remains unreliable in practice due to these magnetic disturbances. \cite{calibration} tackles this issue by developing an iterative calibration method for inertial and magnetic sensors. While \cite{Fan2018} highlights how magnetic interference influences attitude estimation and discusses methods for decoupling attitude (pitch and roll) estimates from magnetometer disturbances. \cite{martin2007invariant} and \cite{Chen2023} construct an additional inertial vector to decouple attitude from magnetic measurements. The technique demonstrated in \cite{Chen2023} shows promising results in improving quaternion EKF performance under practical conditions and compares favourably against other estimators \cite{Madgwick2020}, \cite{Wu2016}, \cite{Suh2020}. 
% An invariant nonlinear observer developed in \cite{martin2007invariant}, uses an additional inertial vector, derived as the cross-product of gravitational acceleration and the Earth's magnetic field. This approach successfully decouples attitude from magnetic measurements. Similarly, recent work by \cite{Chen2023} constructs a second reference vector by removing the component of the magnetometer reading along the accelerometer reading. This method demonstrates promising results in improving quaternion EKF performance under practical conditions and compares favorably against other estimators \cite{Madgwick2020}, \cite{Wu2016}, \cite{Suh2020}.

%An Extended Complementary Filter for Full-Body MARG Orientation Estimation \cite{Madgwick2020}
%Fast Complementary Filter for Attitude Estimation Using Low-Cost MARG Sensors \cite{Wu2016}
%Attitude Estimation Using Inertial and Magnetic Sensors Based on Hybrid Four-Parameter Complementary Filter \cite{Suh2020}

A key observation is that these approaches use the second and third columns of the Tri-Axial Attitude Determination (TRIAD) generated rotation matrix \cite{shuster1981three}. Although TRIAD, constructed from two reference vectors, is known to be suboptimal due to its disregard of some vector information \cite{bar1997optimized}, this characteristic benefits our goal of decoupling magnetometer readings from attitude estimation.
% markley2008optimal
A closely related work uses the TRIAD to provide a coarse 3D orientation estimate, which is then fed to the multiplicative EKF \cite{Kinatas2023}. In this work we simply "pre-process" the magnetometer vector by using the second or third column of the rotation matrix generated by TRIAD as the second reference vector in the manifold EKF. 
% An alternate approach to achieving this could be tuning the Measurement Noise Covariance Matrices but this reduces the responsiveness of the estimator to changes in the yaw axis.

% The organization of this paper is as follows. Section 2 reviews the Manifold EKF as described in \cite{bernal2019kalman}. Section 3 details the extension and modifications applied to the estimation algorithms. Section 4 presents experimental results, and Section 5 concludes the paper.
This paper reviews the Manifold EKF as described in \cite{bernal2019kalman} in Section 2. Section 3 details the extensions and modifications applied to the estimation algorithms. Section 4 presents experimental results, and Section 5 concludes the paper.

% We marry the manifold ekf with the disturbance rejection
%%%

%%%%%%%%%%%%%%%%%%%%%%%%%%%%%%%%%%%%%%%%%%%%%%%%%%%%%%%%%%%%%
%%%%%%%%%%%%%%%%%%%%%%%%%%%%%%%%%%%%%%%%%%%%%%%%%%%%%%%%%%%%%
\section{Manifold EKF Description}
% \todo{}
\noindent This section reviews the Manifold EKF as described in \cite{bernal2019kalman}.

%%%%%%%%%%%%%%%%%%%%%%%%%%%%%%%%%%%%%%%%%%%%%%%%%%%%%%%%%%%%%
\subsection{Preliminaries} \label{sub:preliminaries}
A unit quaternion  $ \in \mathsf{S}^3$ (a manifold) is used to represent orientation. A quaternion can be represented using several notations: $\bm{q} = q_0 + q_1 \bm{i} + q_2 \bm{j} + q_3 \bm{k} \equiv (q_0, q_1, q_2, q_3)^T \equiv (q_0, \mathbf{q})^T$, where $\mathbf{q} = (q_1, q_2, q_3) \in \mathbb{R}^3$. To define probability distributions and their evolution using the Kalman filter, it is essential to translate to Euclidean space, and hence there is a need to define charts for the manifold. \cite{Markley2003} discusses the three-component "attitude error representations". Viewing the quaternion as a manifold these "attitude-error representations" are simply different charts $\varphi$ that map a point in the manifold to a point $\bm{e}$ in $\mathbb{R}^3$. Four charts: Orthographic, the Rodrigues Parameters, the Modified Rodrigues Parameters, and the Rotation Vector; are summarized in \cite{bernal2019kalman}.

% $$\mathbb{R}^3 \ni \bm{e}=\varphi(\bm{q})$$

% \todo
% \begin{table}[h!]
% \centering
% {\def\arraystretch{1.75}
% \begin{tabular}{m{1.5cm} m{1.75cm} m{4cm} }
%  \hline
%  Chart & $\bm{e}=\varphi(\bm{q})$ & $\bm{q}=\varphi^{-1}(\bm{e})$ \\
%  \hline
%  Orthographic   & $2 \mathbf{q}$  & $\begin{pmatrix}\sqrt{1-\frac{||\bm{e}||^2}{4}} & \bm{e}^T/2\end{pmatrix}^T$  \\
 
%  Rodrigues Parameters  & $2 \frac{\mathbf{q}}{q_0}$ & 
%  $\frac{1}{\sqrt{4+||\bm{e}||^2}} \begin{pmatrix} 2 & \bm{e}^T \end{pmatrix}^T$  \\
 
%  Modified Rodrigues Parameters & $4 \frac{\mathbf{q}}{1+q_0}$ & 
%  $\frac{1}{16+||\bm{e}||^2} \begin{pmatrix} 16 + ||\bm{e}||^2 & 8 \bm{e}^T \end{pmatrix}^T$ \\
 
%  Rotation Vector  & $2 \hat{\mathbf{q}} \mathrm{arcsin}(||\mathbf{q}||)$ & 
%  $\begin{pmatrix} cos(\frac{||e||}{2}) & \hat{\bm{e}}^T sin(\frac{||e||}{2})\end{pmatrix}$  \\
% \hline
% \end{tabular}
% }
% \label{table:1}\caption{Chart Definitions.}
% \end{table}

% \mathsf{S}^3 \ni  = \varphi^{-1}(\bm{e})
% All four charts can be approximated by the same second-order approximation for a point $\bm{e} \in \mathbb{R}^3$ near the origin, to a quaternion $\bm{q} \approx (1 - ||\bm{e}||^2/8, \bm{e}/2)^T.$ 
Each chart $\varphi$ introduces a deformation of the space, yet for quaternions near the identity quaternion, the charts act approximately as identity transformations. This characteristic is useful for the Kalman Filter, as it means that the space around the identity quaternion is "nearly flat." But this is true only in the neighborhood of the identity quaternion. This property is extended to any unit quaternion $\bm{\bar{q}}$ by expressing a unit quaternion $\bm{q}$ as a "deviation" from $\bm{\bar{q}}$ through the following expression, $\bm{q} = \bm{\bar{q}} \ast \bm{\delta^{\bar{q}}_q}$, where $\ast$ represents quaternion product. A chart $\varphi_{\bar{\bm{q}}}$ 
% for $\bar{\bm{q}}$ 
is defined as $\bm{e}^{\bm{\bar{q}}}_{\bm{q}} = \varphi_{\bm{\bar{q}}}(\bm{q}) = \varphi(\bm{\delta}_{\bm{q}}^{\bm{\bar{q}}})$.
% \todo{} \mathbb{R}^3 \ni 

The Euclidean space associated with the chart $\varphi_{\bm{\bar{q}}}$ is referred to as $\bm{\bar{q}}$-centered chart. Thus $\varphi_{\bm{\bar{q}}}^{-1}$ takes a point $\bm{e}^{\bar{\bm{q}}}_{\bm{q}}$ in $\bar{\bm{q}}$-centered chart and maps it to $\bm{q}$ in the manifold through
\vspace{-10pt}
\begin{equation} \label{eq:q_centered_chart_inv}
    \bm{q} = \varphi_{\bm{\bar{q}}}^{-1}(\bm{e}^{\bar{\bm{q}}}_{\bm{q}}) = \bm{\bar{q}} \ast \varphi^{-1}(\bm{e}^{\bm{\bar{q}}}_{\bm{q}}) = \bm{\bar{q}} \ast \bm{\delta^{\bar{q}}_q}
\end{equation}
\begin{figure}
    \centering
    \includegraphics[width=.75\linewidth]{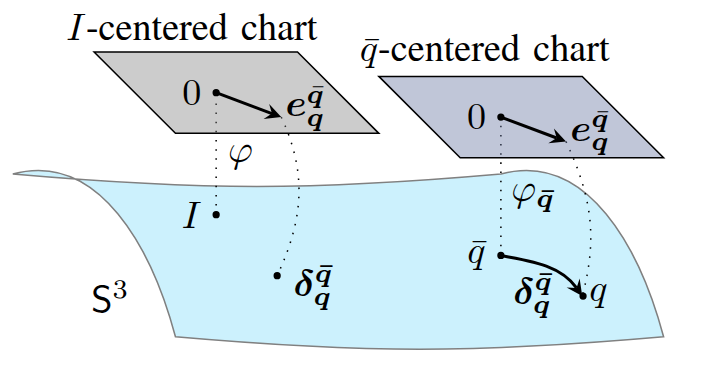}
    \caption{Charts and Deviation}
    \label{fig:charts_and_deviation}
    \vspace{-18pt}
\end{figure}
\vspace{-14pt}

We visualize the notations and concepts introduced above in Fig. \ref{fig:charts_and_deviation}. The distribution is defined in the $\bar{\bm{q}}$-centered chart:
\vspace{-16pt}
\begin{center}
% \begin{subequations} \label{eq:distribution}
    % \begin{equation} 
    % \label{eq:distribution}
    % \label{eq:expectation}
    $ \bar{\bm{e}}^{\bar{\bm{q}}}_{\bm{q}} \triangleq E[\bm{e}^{\bar{\bm{q}}}_{\bm{q}}]
    % \end{equation}
    % \begin{equation} \label{eq:cov_mat}
    \text{ and }
        \mathbf{P}^{\bar{\bm{q}}} \triangleq E[(\bm{e}^{\bar{\bm{q}}}_{\bm{q}}-\bar{\bm{e}}^{\bar{\bm{q}}}_{\bm{q}})(\bm{e}^{\bar{\bm{q}}}_{\bm{q}}-\bar{\bm{e}}^{\bar{\bm{q}}}_{\bm{q}})^T]
    $
\end{center}
%%%%%%%%%%%%%%%%%%%%%%%%%%%%%%%%%%%%%%%%%%%%%%%%%%%%%%%%%%%%%
%
%
\subsection{Motion Equations and Measurement Model}
%
% \todo[inline,color=cyan]{The notation in this section is very confusing.
% You have a subscript $t$ that can be got rid of - it does not make any sense. }
% %
% \red{I would use this:
% %
% \begin{subequations}  \label{eq:motn_eqn}
%     \begin{equation} \label{eq:motn_eqn_omega}
%         \frac{d\bm{\omega}}{dt} = \mathbf{w}^{\bm{\omega}}
%     \end{equation}
%     \begin{equation} \label{eq:motn_eqn_quaternion}
%         \frac{d\bm{q}}{dt} = \frac{1}{2}\bm{q} \ast \begin{pmatrix}0 \\\bm{\omega}\end{pmatrix}    
%     \end{equation}    
% \end{subequations}
% %
% }
%
The complete state of the system at time $t$ is defined by the two-tuple -
$(\bm{q}_t, \bm{\omega}_t) $ - with the orientation $\bm{q}_t \in \mathsf{S}^3$ (unit quaternion) and the angular velocity in body frame $\bm{\omega}_t$ in rad/sec. The following motion equations are used for the prediction step of the Kalman Filter:
\vspace{-4pt}
%
% \blue{I would get rid of this:
%
% \begin{subequations}  \label{eq:motn_eqn}
    % \begin{equation} \label{eq:motn_eqn_omega}
    %     \frac{d\bm{\omega}(t)}{dt} = \mathbf{w}^{\bm{\omega}}_t
    % \end{equation}
    % \begin{equation} \label{eq:motn_eqn_quaternion}
    %     \frac{d\bm{q}(t)}{dt} = \frac{1}{2}\bm{q}(t) \ast \begin{pmatrix}0 \\\bm{\omega}(t)\end{pmatrix}    
    % \end{equation}   
    \begin{equation} 
    \label{eq:motn_eqn}
    % \label{eq:motn_eqn_omega}
        \frac{d\bm{\omega}}{dt} = \mathbf{w}^{\bm{\omega}}
    % \end{equation}
    \text{ and }
    % \begin{equation} \label{eq:motn_eqn_quaternion}
        \frac{d\bm{q}}{dt} = \frac{1}{2}\bm{q} \ast \begin{pmatrix}0 \\\bm{\omega}\end{pmatrix}    
    \end{equation} 
% \end{subequations}
%
% }
\vspace{-12pt}

\noindent $\bm{\mathbf{w}}^{\bm{\omega}}$ is process noise 
% with covariance $Q_\omega$
: $\mathbf{w}^{\bm{\omega}}(t) \sim \mathcal{N}(0,Q_\omega)$.

We rely on two sensors for measurements: a gyroscope and an accelerometer. The gyroscope measures the angular velocity $\bm{\omega}$ in the body frame . We assume the gyroscope is calibrated to ensure that the biases are zero. The accelerometer measures acceleration in the body frame, so when the body is not accelerating it measures the gravity vector. We do not include the disturbances due to acceleration of the body in the measurement model.
%
%
% \red{Note the changes again:
%
The measurement model relates measurements $\bm{z}_{\bm{a}}$ and $\bm{z}_{\bm{\omega}}$ with the states $\bm{q}$ and $\bm{\omega}$ 
%
%\footnote{external disturbances are not included in the accelerometer measurement model}
%
% We know that the accelerometer measurement $\bm{a}_t$ in the external reference frame must measure $\bm{a}_r := -\bm{g}$.
%
% We assume we have sensors giving measurements of angular velocity $\bm{\omega}_t$ and of a reference vector $\bm{v}_t$ whose value $\bm{v}_r$ (r for reference) expressed in the external reference frame is known. 
%
% \begin{subequations} \label{eq:meas_mod}
\vspace{-4pt}
    \begin{equation}
    \label{eq:meas_mod}
        \bm{z}_{\bm{a}} = \mathbf{R}^T_{\bm{q}} \bm{a}_r + \mathbf{v}^a
    % \end{equation}
    \text{ and }
    % \begin{equation}
        \bm{z}_{\bm{\omega}} = \bm{\omega} + \mathbf{v}^\omega,
    \end{equation}
% \end{subequations}
%
%
\noindent where $\mathbf{R}^T_{\bm{q}}$ is the transpose of the rotation matrix corresponding to the quaternion $\bm{q}$. Since we are only interested in the direction of the accelerometer reading, $\bm{z}_{\bm{a}}$ is actually the normalized accelerometer reading. $\bm{a}_r$ is the normalized gravitational acceleration vector in the global frame which is defined to be $(0, 0, -1)^T$, in the North-East-Down (NED) frame convention 
% \footnote{In the experiments $\bm{a}_r$ is found during calibration, by taking a normalized average of the readings taken keeping the body at rest at identity orientation.}.
$\mathbf{v}^\omega$ and $\mathbf{v}^a$ are measurement noise with zero mean and covariance $R_a$ $R_\omega$.
%
% }
% 
%%%%%%%%%%%%%%%%%%%%%%%%%%%%%%%%%%%%%%%%%%%%%%%%%%%%%%%%%%%%%

\subsection{Attitude Estimator Description - Manifold EKF}

% Our knowledge of the state at a time $t$, having included measurements up to a time $t_n$, is described by a distribution encoded in the collection of mathematical objects 
% $$(\varphi, \bar{\bm{q}}, \bar{\bm{x}}_{t|t_ n}^{\bar{\bm{q}}}, \mathbf{P}_{t|t_n}^{\bar{\bm{q}}})$$ 
% $$\begin{pmatrix}
    % \varphi, & \bar{\bm{q}}, & \bar{\bm{x}}_{t|t_ n}^{\bar{\bm{q}}}, & \mathbf{P}_{t|t_n}^{\bar{\bm{q}}}
% \end{pmatrix}$$
% as described in \ref{sub:preliminaries}. $\bar{\bm{x}}_{t|t_ n}^{\bar{\bm{q}}} = (\bar{\bm{e}}_{\bar{\bm{q}}_{t|t_ n}}^{\bar{\bm{q}}}, \bar{\bm{\omega}}_{t|t_n})$ is the expected value of the distribution, and $\mathbf{P}_{t|t_n}^{\bar{\bm{q}}} = E[(\bm{x}^{\bar{\bm{q}}}_{t|t_n}-\bar{\bm{x}}^{\bar{\bm{q}}}_{t|t_n})(\bm{x}^{\bar{\bm{q}}}_{t|t_n}-\bar{\bm{x}}^{\bar{\bm{q}}}_{t|t_n})^T]$ is its $6\times6$ covariance matrix, both expressing the quaternion distribution in the $\bar{\bm{q}}$-centered chart. 

% \blue{
The discrete-time Kalman Filter involves a discrete system with a sampling interval $T$. We denote the discrete stages with the notation $n$. 
% Henceforth, we shall use only the discrete index $n$.
%
% }
Having included measurement up to a time $(n-1)T$, the distribution is expressed in the $\Bar{\bm{q}}_{n-1|n-1}$ centred chart to minimize the deformation caused by the chart.  This is encoded in the collection of the four initial mathematical objects: 
\vspace{-4pt}
$$\begin{pmatrix}
    \varphi, & \bar{\bm{q}}_{n-1|n-1}, & \bar{\bm{x}}_{n-1|n-1}^{\bar{\bm{q}}_{n-1|n-1}}, & \mathbf{P}_{n-1|n-1}^{\bar{\bm{q}}_{n-1|n-1}}
\end{pmatrix}.$$
% =   \begin{pmatrix}
%                 0\\ 
%                 \bar{\bm{\omega}}_{n-1|n-1}
            % \end{pmatrix}
% $\Bar{\bm{q}}_{n-1|n-1}$-centered chart is used to represent the distribution, to minimize the deformation caused by the chart. 
% This distribution has a mean
% $\bar{\bm{x}}_{n-1|n-1}^{\bar{\bm{q}}_{n-1|n-1}} 
%         =   \begin{pmatrix}
%                 \bar{\bm{e}}_{\bar{\bm{q}}_{n-1|n-1}}^{\bar{\bm{q}}_{n-1|n-1}} = \bm{0}& 
%                 \bar{\bm{\omega}}_{n-1|n-1}
%             \end{pmatrix}^T$
% and 6x6 covariance matrix $\mathbf{P}_{n-1|n-1}^{\bar{\bm{q}}_{n-1|n-1}} = E[({\bm{x}}_{n-1|n-1}^{\bar{\bm{q}}_{n-1|n-1}} - \bar{\bm{x}}_{n-1|n-1}^{\bar{\bm{q}}_{n-1|n-1}})({\bm{x}}_{n-1|n-1}^{\bar{\bm{q}}_{n-1|n-1}} - \bar{\bm{x}}_{n-1|n-1}^{\bar{\bm{q}}_{n-1|n-1}})^T]$. 

% $(\phi, \Bar{p}, \bar{x}_{t|n}^{\bar{p}}, P_{t|n}^{\bar{p}})$. 
% is the expected value and $P_{t|n}^{\bar{p}}$ is the 6x6 covariance matrix, both expressed in $\bar{p}$-centered chart.

% $$(\phi, \bar{q}_{n-1|n-1},\bar{\omega}'_{n-1|n-1}, P_{n-1|n-1}^{\bar{q}_{n-1|n-1}})$$
% \subsubsection{Prediction}
%%%%%%%%%%%%%%%%%%%%%%%%%%%%%%%%%%%%%%%%%%%%%%
% \vspace{8pt}
\noindent{Prediction}
\vspace{1pt}
\hrule
\vspace{8pt}

The motion equations \ref{eq:motn_eqn} are used to predict the values of the random variables that describe the state.

\vspace{6pt}

\noindent \textit{\textbf{Predicted state estimate}}:
% \vspace{3pt}
% \begin{subequations}  \label{eq:prediction}

% \begin{equation} \label{eq:pred_q}
    % \boxed{
        $\bar{\bm{q}}_{n|n-1} = \bar{\bm{q}}_{n-1|n-1} \ast \bm{\delta}_{\bar{\bm{q}}_{n|n-1}}^{\bar{\bm{q}}_{n-1|n-1}}
    % }
% \end{equation}
% \begin{equation} \label{eq:pred_w}
    % \boxed{
        \text{ and }
        \bar{\bm{\omega}}_{n|n-1} = \bar{\bm{\omega}}_{n-1|n-1}$
    % }
% \end{equation}

\vspace{3pt}
\noindent This is a consequence of our motion model \ref{eq:motn_eqn} assuming zero angular acceleration. 
% \begin{figure}
%     \centering
%     \includegraphics[width=\linewidth]{Figures/omega_on_KF_timeline.jpg}
%     \caption{$\bm{\omega}$ prediction and update visualized on Kalman Filter time line}
%     \label{fig:pred_correct_omega}
% \end{figure}
% \RS{To clarify the assumption on angular velocity, the evolution of $\bm{\omega}$ is shown on the timeline of the Kalman Filter in Fig. \ref{fig:pred_correct_omega}.} 
% \todo{}
This results in a quaternion "deviation" from $\bar{\bm{q}}_{n-1|n-1}$ to $\bar{\bm{q}}_{n|n-1}$,\\ 
% caused by $\bar{\bm{\omega}}_{n|n-1}$ in $\Delta t_n$ to be:\\
% \begin{equation} \label{eq:pred_del}
    $\bm{\delta}_{\bar{\bm{q}}_{n|n-1}}^{\bar{\bm{q}}_{n-1|n-1}} = \begin{pmatrix}
                            \cos{(\frac{||\bar{\bm{\omega}}_{n|n-1}||\Delta t_n}{2})}\\
                            \frac{\bar{\bm{\omega}}_{n|n-1}}{||\bar{\bm{\omega}}_{n|n-1}||}\sin(\frac{||\bar{\bm{\omega}}_{n|n-1}||\Delta t_n}{2})
                        \end{pmatrix}$
% \end{equation}

\vspace{6pt}

\noindent \textit{\textbf{Predicted error covariance}}:
\vspace{-4pt}
% \begin{equation} \label{eq:pred_P}
    % \boxed{
\begin{center}
        $\mathbf{P}_{n|n-1}^{\bar{q}_{n|n-1}} = \mathbf{F}_n[\mathbf{P}_{n-1|n-1}^{\bar{q}_{n-1|n-1}} + \mathbf{Q}_n]\mathbf{F}_n^T$
\end{center}
    % }
% \end{equation}
\vspace{-4pt}
\noindent where, linearization of the non-linear motion model $\mathbf{F_n}$ and the effect of the process noise on the states $\mathbf{Q}_n$ are:\\
% \begin{equation} \label{eq:pred_F}
    % $\mathbf{F_n} = \begin{bmatrix}
    %         \mathbf{R}^T_{\bm{\delta}_n^{\bar{\bm{q}}_{n-1|n-1}}} & I\Delta t_n\\
    %         0 & I
    %     \end{bmatrix}$
        $\mathbf{F_n}=\begin{bmatrix}
            \mathbf{R}^T_{\bm{\delta}_n} & I\Delta t_n\\
            0 & I
        \end{bmatrix}
        \text{, }
% \end{equation}
% \end{subequations}
% \begin{equation*} \label{eq:pred_Qn}
    \mathbf{Q}_n = \begin{bmatrix}
        Q_\omega \frac{(\Delta t_n)^3}{3}  & -Q_\omega \frac{(\Delta t_n)^2}{2} \\
       -Q_\omega \frac{(\Delta t_n)^2}{2}  &  Q_\omega \Delta t_n
    \end{bmatrix}$.
% \end{equation*}

\noindent Thus our predicted four:
\vspace{-4pt}
$$\begin{pmatrix}
    \varphi, \bar{\bm{q}}_{n|n-1}, \bar{\bm{x}}_{n|n-1}^{\bar{\bm{q}}_{n|n-1}}=   (
                \bar{\bm{e}}_{\bar{\bm{q}}_{n|n-1}}^{\bar{\bm{q}}_{n|n-1}} = \bm{0},  
                \bar{\bm{\omega}}_{n|n-1}
            )^T,  \mathbf{P}_{n|n-1}^{\bar{\bm{q}}_{n|n-1}}
\end{pmatrix}$$

\noindent \textit{\textbf{Measurement Prediction}}

% \begin{subequations}  \label{eq:meas_pred}
Given the predicted state and the measurement model, we predict the measurements,         $\bar{\bm{z}}_{n|n-1}  = \begin{pmatrix}\bar{\bm{z}}_{a_{n|n-1}}, \bar{\bm{z}}_{\omega_{n|n-1}} \end{pmatrix}^T$:
\vspace{-4pt}
\begin{equation} \label{eq:meas_pred}
% \label{eq:meas_pred_a}
    \bar{\bm{z}}_{a_{n|n-1}} = \mathbf{R}^T_{\bm{q}_{n|n-1}}\mathbf{\bm{a}_r}
% \end{equation}
    \text{ and }
% \begin{equation} \label{eq:meas_pred_w}
    \bar{\bm{z}}_{\omega_{n|n-1}}  = \bar{\bm{\omega}}_{n|n-1}.
\end{equation}
% \begin{equation} \label{eq:meas_pred_z}
    % \boxed{

    % }
% \end{equation}
% \end{subequations}

%%%%%%%%%%%%%%%%%%%%%%%%%%%%%%%%%%%%%%%%%%%%%%

\vspace{4pt}
\noindent{Correction/Update}
\vspace{1pt}
\hrule
\vspace{4pt}

% \todo{}
%
% \red{Have made changes in notation here as well:
%
\noindent A \textbf{measurement} is made at time $nT$:
% }
%
% \begin{equation} \label{eq:update_z}
$\bm{z}_n = \begin{pmatrix}
                {\bm{z}}_{a_{n}}&
                {\bm{z}}_{\omega_{n}}
             \end{pmatrix}^T$
% \end{equation}

% \begin{subequations}
\noindent\textit{\textbf{Innovation residual}}:
% \begin{equation}
    $\tilde{\bm{y}}_n = \bm{z}_n - \bar{\bm{z}}_{n|n-1}$
% \end{equation}

\noindent\textit{\textbf{Innovation covariance}}:
% \begin{equation} \label{eq:innov_S}
    % \boxed{
        $\mathbf{S}_{n} = \mathbf{H}_n \mathbf{P}_{n|n-1}^{\bar{\bm{q}}_{n|n-1}} \mathbf{H}_n^T 
                + \begin{bmatrix}
                    R_a & 0\\
                    0 & R_\omega
                \end{bmatrix}$     
    % }
% \end{equation}
where the linearization of the measurement model gives
% \begin{equation} \label{eq:innov_H}
    $\mathbf{H}_n = \begin{bmatrix}
                    [{\bm{z}}_{a_{n}}]_\times & 0\\
                    0 & I
                \end{bmatrix}$
% \end{equation}
where $[\bm{v}]_\times = \begin{bmatrix}
                            0 & -v_3 & v_2\\
                            v_3 & 0 & -v_1\\
                            -v_2 & v_1 & 0
                        \end{bmatrix}.$

% \end{subequations}
% We find the Kalman gain and perform the correction/update.
\noindent \textit{\textbf{Kalman gain}}:
% \begin{equation} \label{eq:kalman_gain}
    $\mathbf{K}_n = \mathbf{P}_{n|n-1}^{\bar{\bm{q}}_{n|n-1}} \mathbf{H}_n^T \mathbf{S}_{n}^{-1}$
% \end{equation}

\noindent\textit{\textbf{Corrected state estimate}} in the $\bar{\bm{q}}_{n|n-1}$-centered chart:
\vspace{-4pt}
\begin{equation} \label{eq:update_x}
    \bar{\bm{x}}_{n|n}^{\bar{\bm{q}}_{n|n-1}} = \bar{\bm{x}}_{n|n-1}^{\bar{\bm{q}}_{n|n-1}} + \mathbf{K}_n \tilde{\bm{y}}_n
\end{equation}
\textit{\textbf{Corrected estimate covariance}}:
\vspace{-4pt}
\begin{equation} \label{eq:update_P}
    \mathbf{P}_{n|n}^{\bar{q}_{n|n-1}} = (I_6 - \mathbf{K}_n \mathbf{H}_n) \mathbf{P}_{n|n-1}^{\bar{q}_{n|n-1}}
\end{equation}
% \begin{subequations}\label{eq:update}
% \end{subequations}
Thus we have the updated four:
\vspace{-4pt}
$$\begin{pmatrix}
    \varphi, & \bar{\bm{q}}_{n|n-1}, & \bar{\bm{x}}_{n|n}^{\bar{\bm{q}}_{n|n-1}}=   (
                \bar{\bm{e}}_{\bar{\bm{q}}_{n|n}}^{\bar{\bm{q}}_{n|n-1}}, 
                \bar{\bm{\omega}}_{n|n}
            )^T, & \mathbf{P}_{n|n}^{\bar{\bm{q}}_{n|n-1}}.
\end{pmatrix}$$
\hrule
%%%%%%%%%%%%%%%%%%%%%%%%%%%%%%%%%%
\vspace{8pt}

Finally, the updated unit quaternion $\bar{\bm{q}}_{n|n}$ is obtained and the mean and covariance are found in the $\bar{\bm{q}}_{n|n}$-centered chart. From \ref{eq:q_centered_chart_inv},
$
\bar{\bm{q}}_{n|n} = \varphi_{\bar{\bm{q}}_{n|n-1}}^{-1}(\bar{\bm{e}}_{\bar{\bm{q}}_{n|n}}^{\bar{\bm{q}}_{n|n-1}}) = \bar{\bm{q}}_{n|n-1} \ast \varphi^{-1} (\bar{\bm{e}}_{\bar{\bm{q}}_{n|n}}^{\bar{\bm{q}}_{n|n-1}}) = \bar{\bm{q}}_{n|n-1} \ast \bar{\bm{\delta}}_{\bar{\bm{q}}_{n|n}}^{\bar{\bm{q}}_{n|n-1}}.
$
\noindent Since the Kalman update can produce any point in the $\bar{\bm{q}}_{n|n-1}$-centered chart it is "saturated" to the closest point contained in the image of each chart. The point $\bar{\bm{e}}_{\bar{\bm{q}}_{n|n}}^{\bar{\bm{q}}_{n|n-1}}$ in the $\bar{\bm{q}}_{n|n-1}$-centered chart is the origin in the $\bar{\bm{q}}_{n|n}$-centered chart.

\cite{bernal2019kalman} also explains the “covariance correction step” or the "chart update" as the update of the covariance matrix to the new chart. This involves expressing the distribution initially defined in the ${\bar{\bm{q}}_{n|n-1}}$-centered chart, represented $\mathbf{P}_{n|n}^{\bar{\bm{q}}_{n|n-1}}$, in the ${\bar{\bm{q}}_{n|n}}$-centered chart, represented $\mathbf{P}_{n|n}^{\bar{\bm{q}}_{n|n}}$.  \cite{ge2023note} also generalises this idea to any other manifold. But we shall skip this step since it has also been shown in \cite{bernal2019kalman} that, in practice, we will obtain essentially the same accuracy in our estimations with or without the covariance correction step. Thus we have have the final four, $\begin{pmatrix}
    \varphi, & \bar{\bm{q}}_{n|n}, & \bar{\bm{x}}_{n|n}^{\bar{\bm{q}}_{n|n}}, & \mathbf{P}_{n|n}^{\bar{\bm{q}}_{n|n}}.
\end{pmatrix}$ which is equivalent to the condition we started the iteration in.

% given the rate at which the estimator runs on a micro-controllers, which also means small $\bm{\delta_n^\omega}$, 

%%%%%%%%%%%%%%%%%%%%%%%%%%%%%%%%%%%%%%%%%%%%%%%%%%%%%%%%%%%%%%%%%

% Extended and Modified Estimator Description
\section{3D Orientation Estimator Description}

The previous section described the machinery and the requisite equation for
the Manifold EKF. However, the accelerometer is insensitive to rotations in the horizontal plane. 
% \magenta{
This can be shown by the fact that, in \ref{eq:meas_mod}, the measurement $\bm{z}_{\bm{a}_t}$ at some orientation $\mathbf{R}_{\bm{q}_t}$ and at another orientation $ \mathbf{R}_z \mathbf{R}_{\bm{q}_t} $ is the same. When $\bm{a}_r = (0,0,1)^T$ and $\mathbf{R}_z$ is a rotation matrix associated with a rotation about the z-axis.
% $\mathbf{R}_z = [\cos{\theta}, -\sin{\theta}, 0; \sin{\theta}, \cos{\theta}, 0; 0, 0, 1]$ and when $\bm{a}_r = (0,0,1)^T$ .
% } 
This makes the accelerometer ineffective in correcting yaw errors. This results in a drift in the estimated orientation about the yaw axis caused by the integration of errors over time. To address this drift, we introduce a second reference vector, the magnetometer. This is described in the
next subsection.
\subsection{Manifold EKF2}

% In general, an estimator using a single reference vector $v_t$ would be prone to drift by a rotation in the axis defined by the reference vector in the external reference frame. This is because since the sensor would be insensitive to .  
The magnetometer measures the magnetic field in the body frame. 
So when the sensor is not influenced by other magnetic fields ("disturbances"), it measures the earth's magnetic field. The measurement model \ref{eq:meas_mod} is now extended to 
incorporate the additional measurement - the magnetometer.
%
% The external disturbance in the measurement is not included for simplicity but can be done ea...
\vspace{-12pt}
\begin{equation}\label{eq:mag}
    \bm{z}_{\bm{m}_t} = \mathbf{R}^T_{\bm{q}_t}\bm{m}_r + \mathbf{r}_t^m
\end{equation}

\vspace{-6pt}
Just as with the accelerometer, since we are only interested in the direction of the magnetometer measurement, $\bm{z}_{\bm{m}_t}$ is the normalized magnetometer measurement. $\bm{m}_r$ is the normalized earth's magnetic field vector in the global frame which can be found from the World Magnetic Model. 
% \magenta{
At the location where the experiments were conducted we find the inclination to be $+28.65$ deg (downward) and declination to be negligible, which would result in $\bm{m}_r = (0.8775, 0, -0.4795)^T$.
% }

Now we restate the equations for the manifold EKF with the additional measurement.
Since the state and the motion equations remain unchanged, the prediction step remains unchanged. The measurement prediction, $\bar{\bm{z}}_{n|n-1} = \begin{pmatrix}
        \bar{\bm{z}}_{\bm{m}_{n|n-1}}, 
        \bar{\bm{z}}_{\bm{a}_{n|n-1}}, 
        \bar{\bm{z}}_{\bm{\omega}_{n|n-1}}
    \end{pmatrix}^T$,
\noindent now in addition to \ref{eq:meas_pred} also includes $\bar{\bm{z}}_{\bm{m}_{n|n-1}} = \mathbf{R}^T_{\bm{q}_{n|n-1}}\bm{m}_r$. The measurement arriving at time $t_n$ is:
$\bm{z}_n = \begin{pmatrix}
                {\bm{z}}_{m_{n}},
                {\bm{z}}_{a_{n}},
                {\bm{z}}_{\omega_{n}}
             \end{pmatrix}^T
$. The innovation covariance is:
    \begin{equation}
        \mathbf{S}_{n|n-1} = \mathbf{H}_n \mathbf{P}_{n|n-1}^{\bar{\bm{q}}_{n|n-1}} \mathbf{H}_n^T + \begin{bmatrix}
                R_m & 0 & 0\\
                0 & R_a & 0\\
                0 & 0 & R_\omega
            \end{bmatrix}
    \end{equation}
where the linearization of the measurement model gives
    \begin{equation}
        \mathbf{H}_n = \begin{bmatrix}
        [\bar{\bm{z}}_{\bm{m}_{n|n-1}}]_\times & 0\\
        [\bar{\bm{z}}_{\bm{a}_{n|n-1}}]_\times & 0\\
        0 & I
    \end{bmatrix}
    \end{equation}
\begin{subequations}\label{eq:mekf2}
\end{subequations}
% $$S_{n|n-1} = H_n P_{n|n-1}^{\bar{q}_{n|n-1}} H_n^T + \begin{bmatrix}
%     R^T(\bar{q}_{n|n-1})Q_n^vR(\bar{q}_{n|n-1})+R_n^v & 0 & 0\\
%     0 & R^T(\bar{q}_{n|n-1})Q_n^vR(\bar{q}_{n|n-1})+R_n^v & 0\\
%     0 & 0 & R_n^\omega
% \end{bmatrix}$$
The Kalman gain and update equations remain unchanged (the dimensions have changed). Obtaining the updated quaternion and the chart update steps also remain unchanged.

% $$\mathbf{K}_n = \mathbf{P}_{n|n-1}^{\bar{\bm{q}}_{n|n-1}} \mathbf{H}_n^T \mathbf{S}_{n|n-1}^{-1}$$
% $$\bar{\bm{x}}_{n|n}^{\bar{\bm{q}}_{n|n-1}} = \bar{\bm{x}}_{n|n-1}^{\bar{\bm{q}}_{n|n-1}} + \mathbf{K}_n(\bm{z}_n - \bar{\bm{z}}_{n|n-1})$$
% $$\mathbf{P}_{n|n}^{\bar{\bm{q}}_{n|n-1}} = (I_6 - \mathbf{K}_n \mathbf{H}_n) \mathbf{P}_{n|n-1}^{\bar{\bm{q}}_{n|n-1}}$$

\subsection{TRIAD aided Manifold EKF2}
The above-prescribed procedure fixes the problem of rotational drift in the yaw axis but introduces another issue of inconsistency between the two reference vectors - 
the gravity vector measurement $\bm{z}_{a_t}$ and the earth's magnetic field measurement $\bm{z}_{m_t}$ - due to noise and disturbance. Due to the inconsistencies there exists no single $\mathbf{R} \in \mathsf{SO(3)}$ such that both $\bm{z}_{\bm{a}_t} = \mathbf{R}^T \bm{a}_r$ and $\bm{z}_{\bm{m}_t} = \mathbf{R}^T \bm{m}_r$ are simultaneously satisfied. In practice, we find that the accelerometer is relatively more accurate and reliable in measuring the gravity vector, however, the magnetometer reading tends to be easily corrupted.
%
% and outputs a bearing which does not point to $\bm{m}_r$ in the inertial frame. 

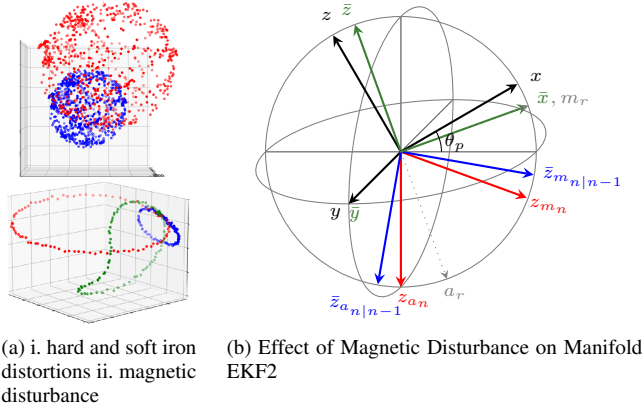
\begin{figure}[h]
  \centering
  % \subfloat[Random sampling (of $\mathsf{S}^2$) of magnetometer readings before calibration (in red) and after calibration (in blue)]{\label{fig:mag_sphere}\includegraphics[width=.4\linewidth]{Figures/mag_calib.png}}
  
  % \qquad
%   \hspace{8pt}
%   \subfloat[magnetometer readings: three standard 360 rotations- yaw pitch and roll.]{\label{fig:b}\includegraphics[width=0.4\linewidth]{Figures/magnetic_disturbance_from_servo.png}}     
%   \caption[The short caption]{Magnetometer Calibration}
%   \label{fig:mag_circles}
% \end{figure}
% \begin{figure}[h]
% \centering
% \subfloat[][Manifold EKF2]{
%     \includegraphics[width=.5\linewidth]{Figures/visualilsing_the_residual.jpg}
%     \label{fig:residual_without_TRIAD}}
% \subfloat[][TRIAD aided Manifold EKF2]{ 
%     \includegraphics[width=.5\linewidth]{Figures/visualilsing_the_effect_of_TRIAD.jpg}
%     \label{fig:residual_with_TRIAD}
% }
\subfloat[i. hard and soft iron distortions ii. magnetic disturbance]{\label{fig:mag_disturb}\includegraphics[width=.3\linewidth]{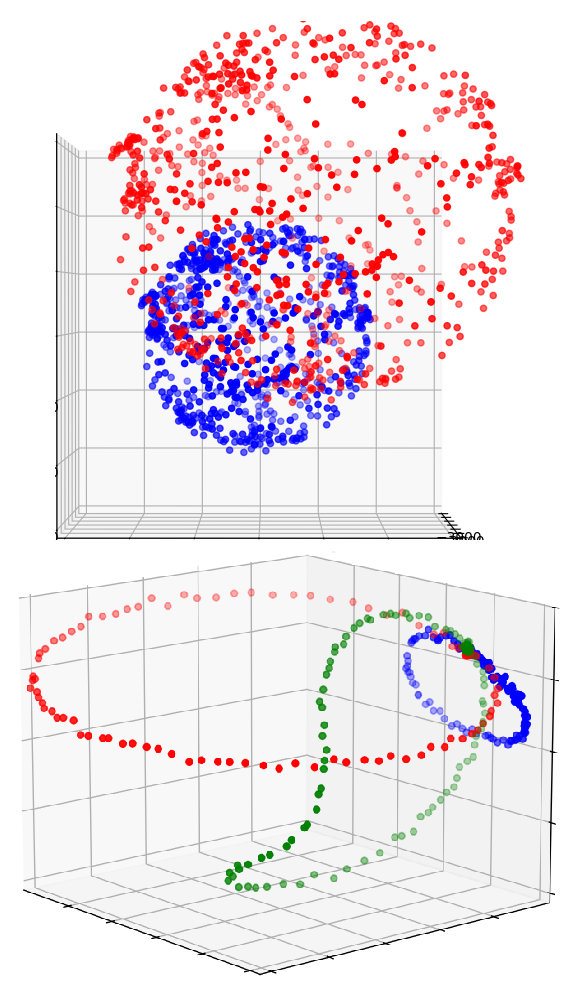}}
    \hspace{8pt}
\subfloat[][Effect of Magnetic Disturbance on Manifold EKF2]{
    \label{fig:residual_without_TRIAD}
\begin{tikzpicture}[scale=1.8]
   \begin{scope}[canvas is zy plane at x=0]
     \draw[gray] (0,0) circle (1cm);
     \draw[gray] (-1,0) -- (1,0) (0,-1) -- (0,1);
   \end{scope}

   \begin{scope}[canvas is zx plane at y=0]
     \draw[gray] (0,0) circle (1cm);
     \draw[gray] (-1,0) -- (1,0) (0,-1) -- (0,1);
     \draw[thick,-stealth] (0,0)--(1,0); %a_r 30deg
     \node at (1.2,-.02) {\scriptsize$y$}; %y
     \node[OliveGreen] at (1.2,.12) {\scriptsize$\bar{y}$}; %y
     
   \end{scope}

   \begin{scope}[canvas is xy plane at z=0]
     \draw[gray] (0,0) circle (1cm);
     \draw[gray] (-1,0) -- (1,0) (0,-1) -- (0,1);
     
     \node[red] at (0.1,-1.1) {\scriptsize$z_{a_n}$};
     \draw[red, thick,-stealth] (0,0)--(0,-1); %z_a_r
    \node[red] at (1.1,-0.4,) {\scriptsize$z_{m_n}$};
     \draw[red, thick,-stealth] (0,0)--(0.93, -0.34); %z_a_r

     \draw[blue, thick,-stealth] (0,0)--(-0.17,-0.98); %a_r 10deg
     \node[blue] at (-.25,-1.15) {\scriptsize$\bar{z}_{a_{n|n-1}}$};%$\mathbf{R}^T_{\bar{q}}a_r$
     \draw[blue, thick,-stealth] (0,0)--(0.98, -0.17); %a_r 10deg
     \node[blue] at (1.35, -.2) {\scriptsize$\bar{z}_{m_{n|n-1}}$};%\mathbf{R}^T_{\bar{q_{n|n-1}}}m_r

     \draw[thick,-stealth] (0,0)--(-0.5,0.86); %z 30deg
     \node at (-0.55,1) {\scriptsize$z$};
     \draw[OliveGreen, thick,-stealth] (0,0)--(-0.34,0.94); %z 20deg
     \node[OliveGreen] at (-0.4,1.05) {\scriptsize$\bar{z}$};
     \draw[gray, dotted,-stealth] (0,0)--(0.34,-0.94); %a_r 20deg
     \node[gray] at (0.4,-1.05) {\scriptsize$a_r$};
     
     \draw[thick,-stealth] (0,0)--(0.86,0.5); %x 30deg
     \node at (1,0.55) {\scriptsize$x$};
     \draw[OliveGreen, thick,-stealth] (0,0)--(0.94,0.34); %x 20deg
     \node[OliveGreen] at (1.05,.4) {\scriptsize$\bar{x}$};
     \draw[gray, dotted,-stealth] (0,0)--(0.94,0.34); %a_r 20deg
     \node[gray] at (1.25,.38) {\scriptsize$, m_r$};
     
     \draw[black] (0.3,0) arc (0:30:.3);
     \node[black] at (0.4,0.07) {\scriptsize$\theta_p$};
   \end{scope}
 \end{tikzpicture}
 }
\caption{Magnetic Disturbances and their effect on the Manifold EKF2.}
\vspace{-16pt}
\end{figure}

% \begin{figure}[h]
% \centering
% \subfloat[][]{
% \subfloat[][magnetometer readings: three standard 360 rotations, before and after calibration]{
% \includegraphics[width=0.4\linewidth]{Figures/mag_measurements.png}
% \label{fig:mag_circles}}
% \qquad
% \subfloat[Subfigure 4 list of figures text][Subfigure 4 caption]{
% \includegraphics[width=0.4\linewidth]{Figures/mag_calib.png}
% \label{fig:subfig4}}
% \caption{Magnetometer Calibration.}
% \label{fig:mag_calib}
% \end{figure}
%
This is reflected in the fact that magnetometers are harder to calibrate and are easily distorted by the effects of soft and hard iron around the sensor. Figure \ref{fig:mag_disturb}(i)  shows a random sample of magnetometer measurements before and after calibration. 
% \magenta{
This distortion in the magnetic field measurements is highly dependent on where the sensor is mounted and needs to be re-calibrated frequently to minimize these distortions. Figure \ref{fig:mag_disturb}(ii) shows magnetometer reading while making a complete rotation in the three standard directions (360$^\circ$ roll, pitch and yaw). The distortion in the green-coloured measurements taken while executing a 360$^\circ$ pitch demonstrates the disturbance caused by the magnets in the servo motor in the experiment setup. This distortion was reduced before conducting the experiments by further increasing the distance between the sensor and the motors.
% }
% \magenta{

Fig. \ref{fig:residual_without_TRIAD} further clarifies the issue of inconsistency between the two reference vectors. It visualises the effect of the magnetic disturbances on the estimated orientation. The figure shows a unit sphere with a black-coloured body frame at the true orientation and a green-coloured body-frame at the estimated orientation. The sphere has 4 "needles" or unit 3-vectors: the predicted measurements $\bar{\bm{z}}_{a_{n|n-1}} = \mathbf{R}^T_{\bm{q}_{n|n-1}}\mathbf{\bm{a}_r}$, $\bar{\bm{z}}_{m_{n|n-1}} = \mathbf{R}^T_{\bm{q}_{n|n-1}}\mathbf{\bm{m}_r}$ in blue, and the measurements $\bm{z}_{a_{n}}$, $\bm{z}_{m_{n}}$ in red. For simplicity in visualisation, the body is only pitched up by $\theta_p$ and we also assume the inclination and declination of the earth's magnetic field is zero. In this visualisation, the accelerometer measures the gravity vector accurately while the magnetometer measurement is off from the North vector by a large margin (in the pitch axis). We see that no orientation simultaneously satisfies both measurements. As a result we observe a large residual vectors $\tilde{\bm{y}}_{a_n} = \bm{z}_{a_n} - \bar{\bm{z}}_{a_{n|n-1}}$ and $\tilde{\bm{y}}_{m_n} = \bm{z}_{m_n} - \bar{\bm{z}}_{m_{n|n-1}}$. This residual can not be made zero and the Kalman Filter converges to an orientation that is "in-between" the two orientations that make either one of the residuals zero.
% } 
% Changing noise characteristics in the measurement model could influence where the orientation converges. One could tune the noise characteristics to improve the estimation but we shall take a different approach that is much simpler in practice.

% The orientation of the cube is \_\_deg when the true rotation is 90deg, which clearly shows the significant effect of this disturbance in the orientation estimated by the Manifold EKF with magnetometer.

% shows a cube oriented at $\bar{\bm{q}}_{n|n}$ viewed from an inertial frame, \ref{fig:su} at identity orientation and \ref{fig:subfig3} after a roll of 90 degrees. 
% \begin{figure}[h]
% \centering
% \subfloat[Subfigure 2 list of figures text][magnetometer reading before and after calibration (with random sweep of the $S^2$)]{
% \includegraphics[width=0.45\linewidth]{Figures/MEKF2_compare.png}
% \label{fig:subfig2}}

% \subfloat[Subfigure 3 list of figures text][magnetometer readings: three standard 360 rotations, before and after calibration]{
% \includegraphics[width=0.45\linewidth]{Figures/MEKF2_compare_2.png}
% \label{fig:subfig3}}

% \qquad
% \subfloat[Subfigure 4 list of figures text][Subfigure 4 caption]{
% \includegraphics[width=0.7\linewidth]{Figures/MEKF2_compare_2.png}
% \label{fig:subfig4}}
% \caption{Magnetometer Calibration.}
% \label{fig:globfig}
% \end{figure}

This issue of magnetometer disturbance affecting the inclination (roll and pitch) estimation has been discussed in reference \cite{Fan2018}. \cite{Fan2018} also discusses different methods of decoupling attitude (roll and pitch) estimation from magnetic disturbance. \cite{martin2007invariant} and  \cite{Chen2023} construct and use a new reference vector in their respective estimators. These new reference vectors are essentially the second and third columns of the TRIAD estimated while using the accelerometer for the first vector and magnetometer for the second vector.

\vspace{3pt}
\noindent \emph{TRIAD}: 
$\mathbf{c}_{1n} = \bm{z}_{\bm{a}_n}$;
$\mathbf{c}_{2n} = \frac{\bm{z}_{\bm{a}_n} \times \bm{z}_{\bm{m}_n}}{|\bm{z}_{\bm{a}_n} \times \bm{z}_{\bm{m}_n}|}$; 
$\mathbf{c}_{3n} = \mathbf{c}_{1n}\times\mathbf{c}_{2n}$;
\begin{center}$\mathbf{R_n}   = \begin{bmatrix}
                \mathbf{c}_{1n} & \mathbf{c}_{2n} & \mathbf{c}_{3n}
                \end{bmatrix}$
\end{center}
                
% \begin{subequations}
% \begin{alignat}{3}
% \mathbf{c}_{1n} & = \bm{z}_{\bm{a}_n}\\
% \mathbf{c}_{2n} & = \frac{\bm{z}_{\bm{a}_n} \times \bm{z}_{\bm{m}_n}}{|\bm{z}_{\bm{a}_n} \times \bm{z}_{\bm{m}_n}|} \\
% \mathbf{c}_{3n} & = \mathbf{c}_1\times\mathbf{c}_2 \\
% \mathbf{R_n}   & = \begin{bmatrix}
%                 \mathbf{c}_{1n} & \mathbf{c}_{2n} & \mathbf{c}_{3n}
%                 \end{bmatrix}
% \end{alignat}
% \end{subequations}
%
% But despite calibration, the inconsistencies between the two reference vectors remain. This can be tackled by choosing the $Q$ and $R$ covariance matrices and therefore trusting the accelerometer more than the magnetometer. But this reduces the response time. Therefore finding the optimal Q and R matrices is difficult in practice.
%
% \magenta{
There is a bias towards the first vector in determining the attitude. The first vector is directly used after normalising, but a component of the second vector is removed while generating the TRIAD estimate. Therefore the first vector is aptly called the anchor \cite{bar1997optimized}. 
This nature of the TRIAD is used to decouple the attitude (roll and pitch) estimation from the magnetic disturbance.
% This nature of the TRIAD works in our favour since using the accelerometer as the anchor is precisely how we decouple the attitude (roll and pitch) estimation from the magnetic disturbance.
% , by choosing the accelerometer as the anchor.
% } 
In the Manifold EKF2 estimator described in the previous section, we
do not employ the magnetometer reading $\bm{z}_{\bm{m}_n}$, instead, we replace it with $\mathbf{c}_{3n}$ from the TRIAD. Also, $\bm{m}_r$ is replaced with $\mathbf{c}_{3r} = \mathbf{a}_r \times \mathbf{c}_{2r}$ where $\mathbf{c}_{2r} = \frac{\bm{a}_r \times \bm{m}_r}{|\bm{a}_r \times \bm{m}_r|}$. Alternatively, this procedure could also be
done with the second column of the TRIAD. This procedure
ensures that the residual in the TRIAD-aided-manifold EKF2 converges
to zero. 

An alternate approach would include the disturbance in the measurement model along with the noise as described in \cite{bernal2019kalman} and then tune the accelerometer and magnetometer noise and disturbance parameter of the estimator. But this would be a tedious task and may result in a slower response time as demonstrated in the experiment section below.

%%%%%%%%%%%%%%%%%%%%%%%%%%%%%%%%%%%%%%%%%%%%%%%%%%%%%%%%

\section{Experimental Setup and Results}
This section presents the experimental setup used for testing and comparing the proposed estimators. It also presents the performance metric used, the results and the observation of the experiments.
\subsection{Experimental Setup}

\begin{figure}[h]
\centering
\includegraphics[width=\linewidth]{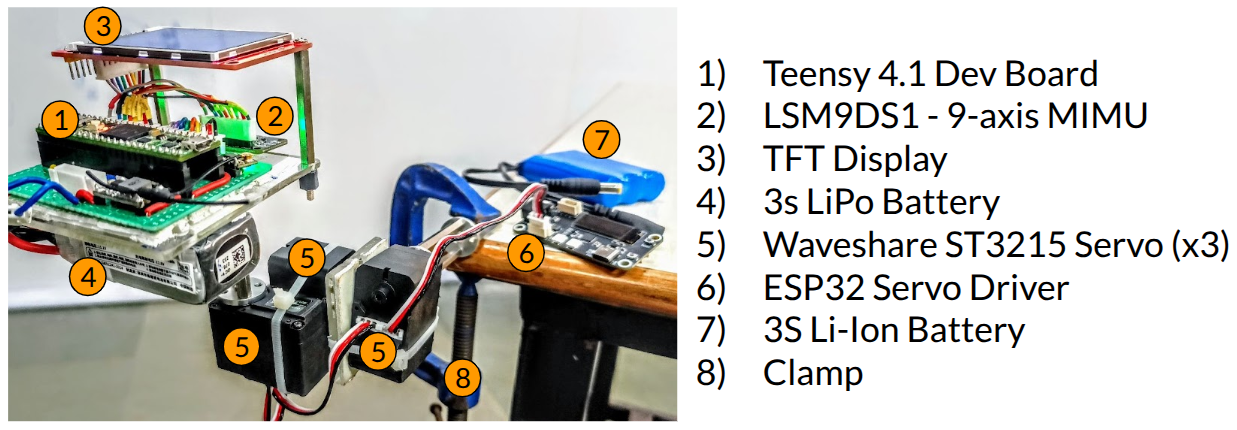}
\caption{Estimator Test Bench.}
\label{fig:est_testbench}
\vspace{-8pt}
\end{figure}

We built two real-world experimental setups to study the effect of the TRIAD aid on the Manifold EKF2 when subjected to real-world magnetic disturbances. However, the challenge with a real-world setup is to get a "true orientation" to compare the accuracy of the estimator. We take an approach similar to \cite{cavallo2014experimental} to obtain the "true orientation". 

Fig. \ref{fig:est_testbench} is a labelled picture of the hardware setup used to compare the estimators. We use a 3-degree-of-freedom (3DOF) robotic arm to orient the sensor at a desired orientation. We use Waveshare ST3215  Servo, which has a position sensor resolution of $360/4096\approx 0.08789$ deg/count for all three joints. We use the ESP32 Servo Driver Expansion Board to interface and control the servo motors.

The estimator is run on the Teensy 4.1 Development Board. The sensor used is the LSM9DS1 (SmartElex Breakout Board). The sensor is interfaced with the microcontroller with SPI at 8MHz clock allowing for a reading time for all nine measurements of just \~26 microseconds. One iteration of the estimators, Manifold EKF2 with and without TRIAD aid, runs in about 78 microseconds. The loop time is not fixed; instead, it is a variable updated using the time measured by the microcontroller. For comparison, we run both estimators in series. The software does not involve any parallel processing, therefore the total loop time is ~182 microseconds while running both estimators. Therefore the loop rate is about 5.5 kHz. But including the time it takes to log the data onto the SD card, the loop rate reduces to about ~270 microseconds or 3.7 kHz.

We set the gyroscope to have a full-scale range of $\pm 500$ dps, output data rates (ODR) of 952 Hz and a 40 Hz low pass filter (LPF) cutoff. 
We set the accelerometer to have a full-scale range of $\pm 16$ g, ODR 952 Hz, 408 Hz anti-aliasing filter bandwidth. 
The magnetometer (LIS3MDL registers) is set to a full-scale range of $\pm 12$ gauss and FAST\_ODR mode allowing it to reach about 1kHz ODR. 
All the above sensor settings are chosen to get the maximum output data rate at the cost of poorer filtering.

Additionally, the setup also includes an onboard display (Color 320x240 TFT) for developer feedback/debugging and visualization of estimator variables. Alternatively, we use the visualization shown in Figure \ref{fig:snapshots} where the data is streamed serially to the laptop and visualized using a Python script. 

\begin{figure}[h]
\centering
\includegraphics[width=\linewidth]{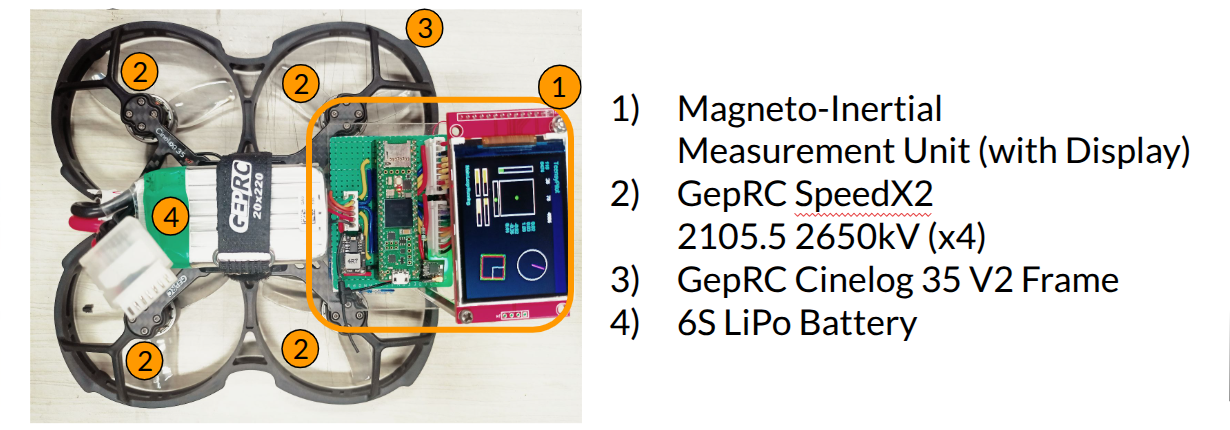}
\caption{Quadcopter UAV hardware setup.}
\label{fig:quadcopter}
\vspace{-8pt}
\end{figure}

We also develop a second experimental setup for testing the estimators under the influences of noise from a real-life scenario of estimating the orientation of a Quadcopter UAV. Fig. \ref{fig:quadcopter} shows an image of the hardware setup for this experiment. In this setup, we place the quadcopter on a flat platform at identity orientation, and introduce measurement noise through motor-induced vibrations and magnetic disturbances, as the motors step through varying RPMs.

The software written for these experiments is part of an open-source drone firmware development effort primarily targeted for the Teensy Development Board. All the code used in these experiments can be found on GitHub under the project named TeensyPilot: https://github.com/arjun-sadananda/TeensyPilot.git. 

% \magenta{
Some points to be noted related to the implementation of the algorithms:
% }
\begin{itemize}
    \item The $\bm{m}_r$ and $\bm{a_r}$ are found during calibration, keeping the body at rest at an orientation that would be defined as the identity orientation. This is done by taking a (normalized) average of 3000 readings for each sensor. 
    \item $Q_\omega$ $R_\omega$ $R_a$ $R_m$ are set as diagonal matrices with all diagonal entries equal. We set $Q_\omega = 10.0 I$, $R_\omega = 0.001 I$, $R_a = 0.01 I$, $R_m = 0.01 I$ for the Manifold EKF2. 
    \item For comparison apart from the Manifold EKF2 with and without TRIAD aid, we also include another Manifold EKF2 where the $R_m$ and $R_a$ covariance matrices are chosen differently to rely more on the accelerometer than the magnetometer. Here we set $R_a = 0.01 I$ and $R_m = 0.05 I$. In the next subsection when we say $R_m > R_a$ we are referring to this version of the Manifold EKF2.
\end{itemize}

\subsection{Experiments and Performance Metric}
\textit{Experiment I}:\\
The Estimator Test Bench is used to make a sequence of rotations as follows: starting at identity;
$90^{\circ}$ yaw; $180^{\circ}$ yaw; identity;
$90^{\circ}$ pitch; $180^{\circ}$ pitch; identity;
$90^{\circ}$ roll; $180^{\circ}$ roll; identity. 
The sensor is held at each orientation for 4 seconds to compare the Manifold EKF2 and TRIAD aided Manifold EKF2 against the expected orientation derived by the kinematics of the servo motor arm. 

\textit{Experiment II}:\\
The drone is kept stationary on a level platform with the propellers removed, and all four motors are stepped through a series of RPMs to induce noise of varying magnitude and frequencies to the various sensors in the Magneto-Inertial Measurement Unit (MIMU). Once again the Manifold EKF2 is compared with and without the TRIAD aid. 

\textit{Performance Metric}\\
To compare the performance of the estimators we shall use rotation error, $\theta_e$ between $\bm{q}$ and $\bm{\bar{q}}$ defined as:
\vspace{-8pt}
$$\theta_e := 2 \arccos[(\bm{\bar{q}}^\ast \ast \bm{q})_0] = 2 \arccos(\bm{\bar{q}} . \bm{q})$$
where $(\bm{\bar{q}}^\ast \ast \bm{q})$ is the deviation between the two quaternions, and the subscript 0 refers to the first component of the quaternion. Also, $\bm{\bar{q}} . \bm{q} >= 0$ is ensured by using the fact that $\bm{q}$ and $-\bm{q}$ represent the same rotation transform.  

% Filter energy is defined as:

% $$\frac{1}{6}{\bm{e}_{\bm{\delta_e}}^I}^T P \bm{e}_{\bm{\delta_e}}^I$$
% $$\frac{1}{6}{\bm{e}_{\bar{\bm{q}}}^{\bm{q}}}^T P \bm{e}_{\bar{\bm{q}}}^{\bm{q}}$$
% $$\frac{1}{6}\bm{x}^T P \bm{x}$$

% The expected value of filter energy is 1, while a smaller or larger value indicates the filter is under-confident or over-confident in its estimation, respectively.

\subsection{Results}
\begin{figure}[h]
\centering
\subfloat[][State Covariance matrices vs time when $R_m = R_a$ and $R_m > R_a$]{
\includegraphics[width=.5\linewidth]{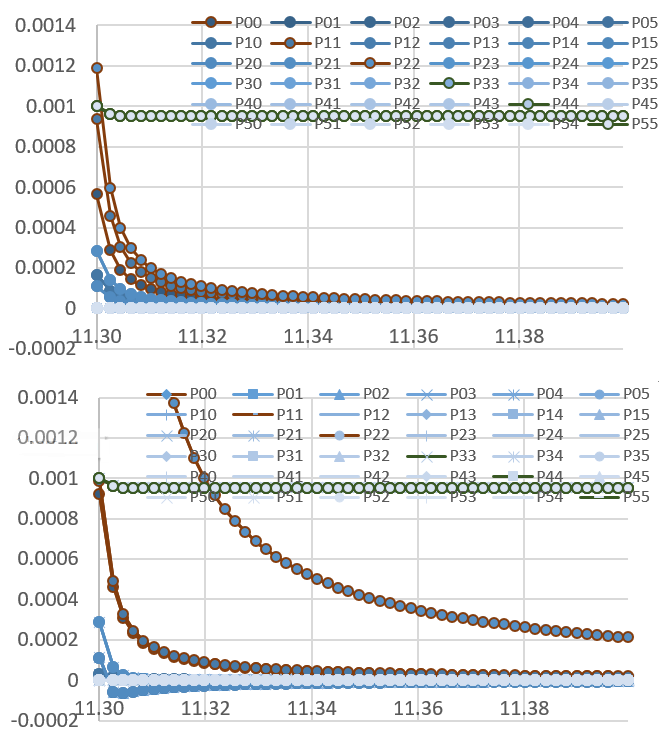}
\label{fig:subfig2}}
% \qquad
\hspace{4pt}
\subfloat[][Projected cubes and 3D compasses visualisation for each of the estimators at $90^\circ$ roll.]{
\includegraphics[width=.4\linewidth]{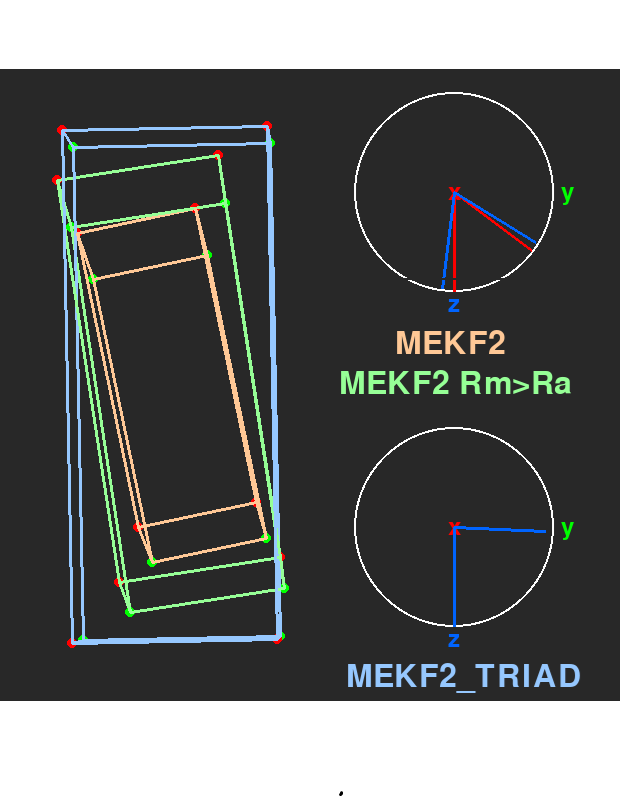}
\label{fig:snapshots}}
\caption{State Covariance evolution and visualisation of the estimators in action}
\label{fig:covariance_vs_time}
\vspace{-8pt}
\end{figure}

Before we look at the results of the experiments we first make an observation comparing the evolution of the state covariance matrices of the three cases. Fig. \ref{fig:covariance_vs_time} shows the components of the state covariance matrix evolving in time for the two cases when $R_m > R_a$ and $R_m = R_a$. As expected we see that it takes longer for the state covariance to converge to the steady state. This indicates that although this estimator would not be affected by magnetic disturbances as much it would reduce the response to yaw rotations.

% \red{above para needs explanation}

\begin{figure}
\centering
  \includegraphics[width=\linewidth]{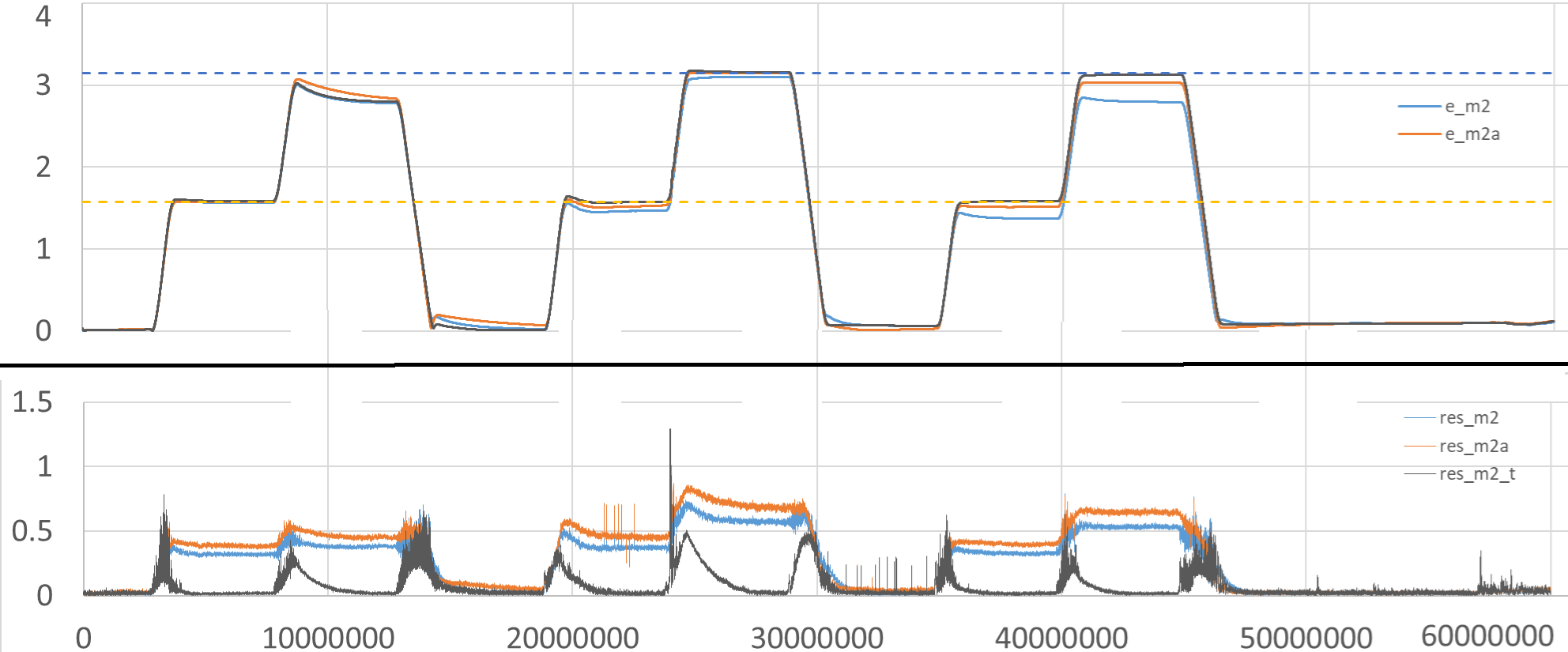}
  \caption{Rotation Error vs time (top) and norm of the residual vs time (bottom) for the three estimators.}
  \label{fig:performance}
  \vspace{-8pt}
\end{figure}

Experiment I clearly demonstrates the superiority of introducing the TRIAD aid to the Manifold EKF2. Fig. \ref{fig:snapshots} shows a snapshot of the experiment after a roll of 90 deg. This figure is a real-world demonstration of what was explained in Fig \ref{fig:residual_without_TRIAD}. The figure also shows virtual cubes at orientation estimated by the respective estimators. The colours of the cube match the colours used for the font in the title. We notice the influence of the magnetometer disturbance in the roll angle estimated. Fig. \ref{fig:performance} quantifies the difference in the performance of the three estimators. Instead of the "rotation error" defined earlier, we plot the "rotation angle" which is calculated by finding the rotation error between the identity quaternion and the estimated quaternion. We know that this rotation angle must be 0, 90 or 180 depending on the different steps of the experiment described in the previous subsection. At a roll of 180$^\circ$ we see an error of 20 deg for the $R_m=R_a$ case, an error of 6.5 deg in the $R_m>R_a$ case and an error of 0.6 deg in the TRIAD aided case. We also note that there is almost no difference between with or without the TRIAD aid in the case of Yaw rotations. Finally, the second plot in Fig \ref{fig:performance} demonstrates how the residual in the Kalman Filter goes to zero in the TRIAD-aided case.

In experiment II we look at the effect of noise in accelerometer and magnetometer measurements on the attitude estimated by the Manifold EKF2 with and without TRIAD  aid. We study two cases when the sensor is soft mounted and when the sensor is rigidly mounted on the quadcopter frame. The major difference between the two cases is that the accelerometer noise is significantly higher than the magnetometer noise in the second case.

\begin{figure}[h]
    \centering
    \subfloat[][soft mount]{
    \includegraphics[width=.5\linewidth]{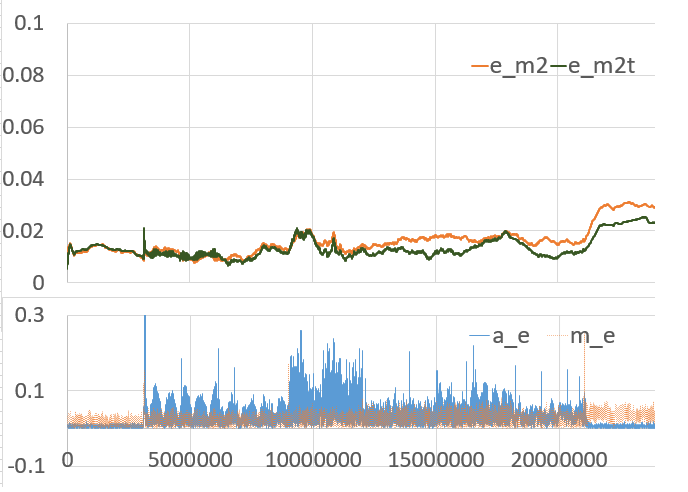}
    \label{fig:soft_mount}}
    % \qquad
    \subfloat[][hard mount]{
    \includegraphics[width=.5\linewidth]{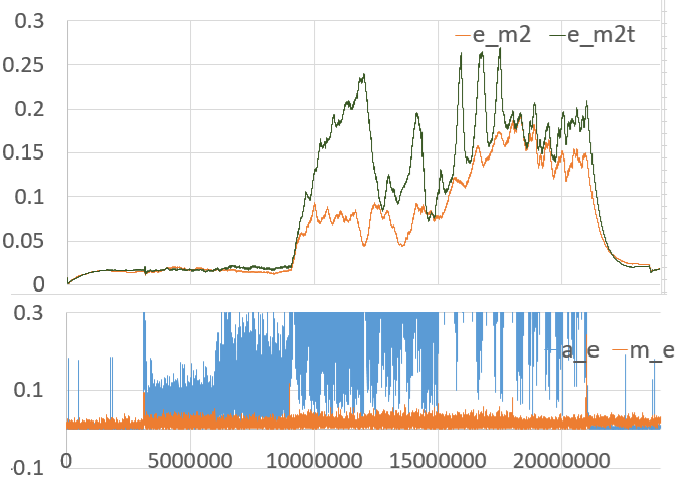}
    \label{fig:hard_mount}}
\caption{Rotation error in Maniold EKF2 and TRIAD aided Maniold EKF2; accelerometer and magnetometer noises.}
\label{fig:quadcopter_noise}
\vspace{-8pt}
\end{figure}
Fig. \ref{fig:quadcopter_noise} shows the rotation error (taking true orientation to be identity) for the two cases. Fig. \ref{fig:quadcopter_noise} also shows the noise in the accelerometer and the magnetometer reading by finding the angle between $\bm{a}_r$,  $\bm{m}_r$ and $\bm{z}_{a_n}$, $\bm{z}_{m_n}$ respectively. We observe that when the noises in the accelerometer and magnetometer are comparable, the TRIAD aid Manifold EKF2 performs better staying close to the true orientation (identity). But when the accelerometer has much larger noise than the magnetometer the TRIAD-aided Manifold EKF2 doesn't help and performs slightly poorly. This is expected because the idea of TRIAD aid is to rely more heavily on the accelerometer by creating a new reference vector that ignores a component of the magnetometer. For the experiment the sensor was mounted rigidly and no filtering was done, which is not the case in practice. However, we do need to note the importance of handling the accelerometer disturbances caused by an accelerating body.
% \todo
\section{Conclusion}

This paper started by reviewing the Manifold EKF, which is an EKF developed using the manifold structure of the quaternion. We extended this work to Manifold EKF2 to include the magnetometer to account for yaw drift. We then discuss in detail the issue of inconsistency between the two reference vectors and how it is resolved by taking the aid of the TRIAD estimator. Finally, we present the results of two experiments that demonstrate how the TRIAD aid improves the performance of the estimator. We also demonstrate what care needs to be taken due to the choice of using the accelerometer as the "anchor" for the TRIAD.

\bibliographystyle{IEEEtran}
\bibliography{IEEEabrv,name1}
\end{document}